\documentclass[11pt]{article}
\usepackage[utf8]{inputenc}
\usepackage{textcomp}
\usepackage{caption}

\usepackage[final]{acl}
\usepackage{float}
\usepackage{placeins}
\usepackage{footnote}
\makesavenoteenv{table}
\usepackage{times}
\usepackage{latexsym}
\usepackage{booktabs}
\usepackage{tabularx}
\usepackage{booktabs}
\usepackage{pifont}
\newcommand{\cmark}{\ding{51}}
\newcommand{\xmark}{\ding{55}}
\usepackage[T1]{fontenc}
\usepackage{subcaption}
\usepackage{textcomp}

\usepackage[utf8]{inputenc}

\usepackage{microtype}
\usepackage{xcolor}
\definecolor{mayank}{RGB}{30, 80, 180}

\usepackage{inconsolata}

\usepackage{textcomp}
\usepackage{amsmath}

\usepackage{tikz}
\usetikzlibrary{positioning, arrows.meta}
\usepackage{graphicx}
\usepackage{pifont}
\renewcommand{\cmark}{\ding{51}}
\renewcommand{\xmark}{\ding{55}}

\title{JobMatchAI \\
\large An Intelligent Job Matching Platform Using Knowledge Graphs, Semantic Search and Explainable AI \\
\vspace{0.5em}
 \small \href{https://mutu.dev/}{Website} \hspace{5 em}
 \small \href{https://github.com/coral-lab-asu/job-hunt-AI.git}{Installation Package} \hspace{5 em} \small \href{https://youtu.be/4jPKvzxZIYU}{Demo Video}}

\author{
  \textbf{Mayank Vyas\textsuperscript{1}},\hspace{1em}
  \textbf{Abhijit Chakraborty\textsuperscript{1}},\hspace{1em}
  \textbf{Vivek Gupta\textsuperscript{1}}
\\\\
  \begin{tabular}{cc}
    \textsuperscript{1}Arizona State University \\
    \texttt{\{mvyas7, achakr40, vgupt140\}@asu.edu}
  \end{tabular}
}

\begin{document}
\maketitle
\begin{abstract}
Recruiters and job seekers rely on search systems to navigate labor markets, making candidate matching engines critical for hiring outcomes. Most systems act as keyword filters, failing to handle skill synonyms and nonlinear careers, resulting in missed candidates and opaque match scores. We introduce JobMatchAI, a production-ready system integrating Transformer embeddings, skill knowledge graphs, and interpretable reranking. Our system optimizes utility across skill fit, experience, location, salary, and company preferences, providing factor-wise explanations through resume-driven search workflows. We release JobSearch-XS benchmark and a hybrid retrieval stack combining BM25, knowledge graph and semantic components to evaluate skill generalization. We assess system performance on JobSearch-XS across retrieval tasks with bootstrap confidence intervals, per-query variance characterization, a skill-extractor impact study, and a directional cross-domain transfer evaluation, and provide a demo video, hosted website, and installable package.
\end{abstract}

\section{Introduction}
Job seekers and recruiters rely on matching systems to navigate increasingly complex labor markets, yet most production platforms still depend on keyword filters enhanced with proprietary heuristics. These systems struggle with skill synonyms (e.g., Kubernetes vs. container orchestration), overlook nonlinear career trajectories, and generate opaque match scores that cannot be audited for compliance with hiring regulations \cite{raghavan_mitigating_2020}. This results in a dual failure: qualified candidates are overlooked, and neither party understands why a particular ranking was produced.

To address this, we introduce JobMatchAI, a production-ready, microservices-based platform for explainable job search and candidate matching. The system addresses three capabilities that are individually studied but rarely integrated into a single deployable stack: (i) hybrid retrieval that combines BM25 lexical search, approximate nearest-neighbor semantic search over Sentence Transformer embeddings, and multi-hop traversal of a skill knowledge graph in Neo4j; (ii) white-box reranking via a multi-factor utility function decomposed into skill fit, experience, location, salary, semantic similarity, and company preference with user-adjustable weights; and (iii) grounded LLM explanations, where a generative model narrates pre-computed scores and knowledge-graph evidence rather than raw documents, ensuring it can explain a ranking but never inflate one.

A key architectural decision sets JobMatchAI apart from prior work: the strict separation of a deterministic scoring layer from a generative explanation layer. Because the LLM receives only the six pre-computed
scores and supporting knowledge-graph paths, not
raw documents, the resulting explanations are
auditable and traceable, a property increasingly
required in hiring contexts \cite{doshi-velez_towards_2017}. Users interact with the system through two modes: Smart Search, where a resume upload triggers profile extraction, knowledge-graph population, and ranked job retrieval; and Keyword Search with Semantic Enrichment, where free-text queries are expanded via graph-based skill associations.
\newline\indent Along with the system, we contribute: (i) A deployable reference architecture integrating semantic parsing, graph storage, lexical indexing, hybrid retrieval, and explainable reranking as stateless microservices. (ii) JobSearch-XS, a skill-grounded benchmark with 1,283 NYC civil-service roles, 30 queries, ${\sim}29$K silver labels, and skill-disjoint train/dev/test splits for evaluating zero-shot skill generalization. (iii) An offline and user evaluation reporting NDCG@10 of 0.81 on JobSearch-XS (7\% over the BM25 baseline at sub-100\,ms latency), complemented by 95\% bootstrap confidence intervals, per-query variance analysis, a skill-extractor impact study, and a cross-domain transfer spot-check, plus a user study with 20 participants assessing relevance, explainability, and usability. (iv) A live web demo and installable package for hands-on exploration.

\section{System Overview}
JobMatchAI is implemented as a set of stateless microservices behind an API gateway, organized along an ingestion $\to$ indexing $\to$ retrieval $\to$ reranking $\to$ explanation pipeline (Figure~\ref{fig:architecture}). The system uses Sentence Transformers (all-MiniLM-L6-v2) for dense encoding, Neo4j for knowledge-graph storage, and Elasticsearch for lexical indexing and approximate nearest-neighbor (ANN) search. The majority of neural computation occurs at ingestion time; online serving involves only vector lookup, graph traversal, and lightweight scoring, keeping median query latency under 100\,ms. The remainder of this section describes each pipeline stage.

\begin{figure}[htbp]
\centering
\resizebox{\columnwidth}{!}{
\begin{tikzpicture}[
    node distance=0.6cm and 1.0cm,
    every node/.style={font=\small},
    stage/.style={
        rectangle, draw=black!70, fill=blue!8,
        minimum width=6.2cm, minimum height=1.1cm,
        align=center, rounded corners=3pt
    },
    retriever/.style={
        rectangle, draw=black!70, fill=green!10,
        minimum width=2.4cm, minimum height=1.4cm,
        align=center, rounded corners=2pt
    },
    arrow/.style={-{Stealth[length=2.5mm]}, thick, draw=black!60},
]

\node[stage] (query) {User Query\\[2pt]\footnotesize\texttt{"Python developer machine learning"}};

\node[stage, below=of query] (enrich) {%
    \textbf{Stage 1: Query Enrichment}\\[2pt]
    \footnotesize Entity Extraction $\bullet$ Skill Extraction $\bullet$ KG Expansion $\bullet$ Embedding};

\node[retriever, below left=1.0cm and -0.8cm of enrich] (lex) {%
    \textbf{Lexical}\\[2pt]\footnotesize ES / BM25\\$k=150$};

\node[retriever, below=1.0cm of enrich] (sem) {%
    \textbf{Semantic}\\[2pt]\footnotesize kNN / HNSW\\$k=150$};

\node[retriever, below right=1.0cm and -0.8cm of enrich] (kg) {%
    \textbf{Knowledge Graph}\\[2pt]\footnotesize Neo4j\\$k=75$};

\node[stage, below=3.4cm of enrich] (fusion) {%
    \textbf{Stage 2: Fusion \& Deduplication}\\[2pt]
    \footnotesize Union by Job ID $\bullet$ RRF / Weighted Fusion $\bullet$ $\max = 400$};

\node[stage, below=of fusion] (filter) {%
    \textbf{Stage 3: Filtering}\\[2pt]
    \footnotesize Query-level constraints $\bullet$ Hard-constraint filter};

\node[stage, below=of filter] (rerank) {%
    \textbf{Stage 4: Personalized Reranking}\\[2pt]
    \footnotesize Skill (0.35) $\bullet$ Experience (0.25) $\bullet$ Location (0.15) $\bullet$ Salary (0.10)};

\node[stage, below=of rerank, fill=orange!15] (results) {\textbf{Ranked Results w/ AI Explanations}};

\draw[arrow] (query) -- (enrich);
\draw[arrow] (enrich) -- (lex.north);
\draw[arrow] (enrich) -- (sem.north);
\draw[arrow] (enrich) -- (kg.north);
\draw[arrow] (lex.south) -- (fusion);
\draw[arrow] (sem.south) -- (fusion);
\draw[arrow] (kg.south) -- (fusion);
\draw[arrow] (fusion) -- (filter);
\draw[arrow] (filter) -- (rerank);
\draw[arrow] (rerank) -- (results);

\node[right=0.2cm of kg, font=\scriptsize\itshape, text=gray, align=left] {Parallel\\(ThreadPool)};
\end{tikzpicture}
}
\caption{End-to-end JobMatchAI hybrid search pipeline architecture.}
\label{fig:architecture}
\end{figure}

\subsection{Ingestion and Semantic Processing}

Crawlers fetch job postings from multiple sources with rate limiting and deduplication; an AI layer standardizes formats before processing.

Documents undergo two parallel processes. Sentences are encoded with \texttt{all-MiniLM-L6-v2} into 384-dimensional embeddings for kNN retrieval. A hybrid extractor using semantic similarity and spaCy patterns identifies skills, locations, companies, and degree requirements. A classifier labels documents as junior, mid-level, or senior.

Both representations are dual-indexed: Elasticsearch receives BM25-indexed text with kNN vectors, while Neo4j stores structured entities as graph nodes. This enables lexical, semantic, and graph-based retrieval from a single ingestion pass. The pipeline supports configurable crawl sources; all evaluation uses the NYC Open Data dataset under open license.
\subsection{Knowledge Graph Design}

The knowledge graph enables JobMatchAI to bridge vocabulary gaps between resumes and job postings through five node types: \texttt{Candidate}, \texttt{Job}, \texttt{Skill}, \texttt{Location}, and \texttt{Company}, connected by relations:

\begin{itemize}    \item \texttt{HAS\_SKILL} and \texttt{REQUIRES\_SKILL} connect candidates and jobs to skills
  (e.g., (Candidate)$\to$[:HAS\_SKILL]$\to$(Python);
  (Job)$\to$[:REQUIRES\_SKILL]$\to$(Python)).
  When both edges exist, the candidate--job pair
  shares a verified skill match.    \item \texttt{RELATED\_TO} edges encode skill associations (e.g., Kubernetes $\leftrightarrow$ Docker), curated from the ESCO v1.1.1 taxonomy augmented with a manually-curated synonym table covering domain-specific gaps (e.g., emerging ML frameworks and tooling vocabulary not yet in ESCO). The full synonym mapping is released alongside the benchmark. \texttt{LOCATED\_IN} links entities to locations.\end{itemize}

Multi-hop traversal over \texttt{RELATED\_TO} edges enables latent skill discovery: candidates skilled in Kubernetes match jobs requiring "container orchestration," unavailable through lexical retrieval alone.

The graph provides interpretable features for the reranker (see Sec~\ref{sec:reranker}): skill set overlap, role distance, and skill-cluster reachability become auditable utility inputs.

\subsection{Hybrid Search Pipeline}\label{sec:hybrid_pipeline}

The hybrid search pipeline is the central technical component. Given a user query which can be either free text or an extracted resume profile, it proceeds in five stages.

\paragraph{Stage 1: Query Enrichment.} The raw query $q$ is expanded into a structured representation $\hat{q} = \langle E, S, S^{+}, \mathbf{e}_q, K \rangle$ containing named entities $E$, extracted skills $S$, graph-expanded skills $S^{+} \supseteq S$ (via depth-2 traversal of \texttt{RELATED\_TO} edges), a dense embedding $\mathbf{e}_q$, and keywords $K$. Unlike pseudo-relevance feedback, this expansion is grounded in the curated knowledge graph, preventing topic drift.

\paragraph{Stage 2: Parallel Multi-Source Retrieval.} Three retrievers execute concurrently via a thread pool:

\begin{itemize}
    \item \textbf{Lexical} (Elasticsearch/BM25, $k=150$): field-boosted keyword search with structured filters.
    \item \textbf{Semantic} (ANN, $k=150$): approximate nearest-neighbor search over pre-computed 384 dimensional embeddings.
    \item \textbf{Graph} (Neo4j Cypher, $k=75$): skill-to-job traversal over \texttt{REQUIRES\_SKILL} edges, seeded by both extracted and graph-expanded skills.
\end{itemize}

Parallel execution keeps latency under 100\,ms while maximizing recall diversity; each channel contributes candidates the others miss.

\paragraph{Stage 3: Query-Adaptive Reciprocal Rank Fusion.} Ranked lists are merged using Reciprocal Rank Fusion (RRF):

\begin{equation}\label{equ:1}
\mathrm{RRF}(d) = \sum_{r \in R} \frac{w_r}{k + \mathrm{rank}_r(d)}, \quad k = 60
\end{equation}

Crucially, the weights $w_r$ are query-adaptive: short queries ($|q| \leq 2$ tokens) prioritize the knowledge graph ($w_{\mathrm{kg}} = 0.7$), while longer queries shift weight toward text matching ($w_{\mathrm{text}} = 0.6$). This adaptive strategy yields consistent gains over fixed-weight baselines across query-length buckets (as discussed in Sec~\ref{sec:evaluation}).

\paragraph{Stage 4: Hard-Constraint Filtering.} Non-negotiable requirements (visa sponsorship, minimum degree, required certifications) are applied as post-retrieval filters, preventing desirable-but-ineligible results from consuming reranking budget.

\paragraph{Stage 5: Multi-Factor Reranking.} Filtered candidates are scored by the utility function detailed in Sec~\ref{sec:reranker}.

\subsection{Explainable Reranking}\label{sec:reranker}

The final ranking is produced by a weighted utility function:

\begin{equation}\label{equ:2}
U(c,j) = \sum_{f \in \mathcal{F}} w_f \cdot \phi_f(c,j)
\end{equation}

where \[\mathcal{F} = \left\{\begin{array}{l}\text{Skill}, \text{Experience}, \text{Location}, \\\text{Salary}, \text{Semantic}, \text{Company}\end{array}\right\}\]
 and each $\phi_f \in [0,1]$ is a normalized factor score. Table~\ref{tab:weights} lists default weights and feature definitions. Users may adjust weights interactively via frontend sliders; the utility function recomputes in real time.
\begin{table}[t]
\centering\small
\caption{Default reranking weights. Users may adjust these interactively; the utility function recomputes in real time.}
\label{tab:weights}
\begin{tabular}{@{}lcc@{}}
\toprule
\textbf{Factor} ($f$) & $w_f$ & \textbf{Feature} ($\phi_f$) \\
\midrule
Skill match       & 0.35 & Jaccard + KG relatedness bonus \\
Experience        & 0.25 & Level-distance penalty \\
Location          & 0.15 & Tiered: exact / state / remote \\
Salary            & 0.10 & Expectation-to-midpoint ratio \\
Semantic sim.     & 0.10 & $\cos(\mathbf{e}_c, \mathbf{e}_j)$ \\
Company fit       & 0.05 & Industry \& size preference \\
\bottomrule
\end{tabular}
\end{table}

A distinguishing design property is the strict separation of scoring from explanation. The utility function is fully white-box: each factor score $\phi_f$ is computed from interpretable inputs such as skill-set overlap or location match tier, with no learned parameters beyond the embeddings.
The six factor scores and their contributing knowledge-graph paths are then passed to an LLM, which synthesizes a cohesive narrative such as:

\begin{quote}
\small
``This role is a strong match (87\%) primarily due to verified expertise in Python and React. The location preference (Remote) conflicts slightly with the on-site requirement, reducing the location score.''
\end{quote}

Because the LLM receives pre-computed scores and graph evidence rather than raw documents, it can explain a ranking but cannot hallucinate a high one. This separation provides the fluency of generative models with the auditability of symbolic scoring, a property critical for hiring compliance.

\begin{figure*}[t]
  \centering
  \begin{subfigure}[b]{0.48\textwidth}
    \centering
    \includegraphics[width=\linewidth]{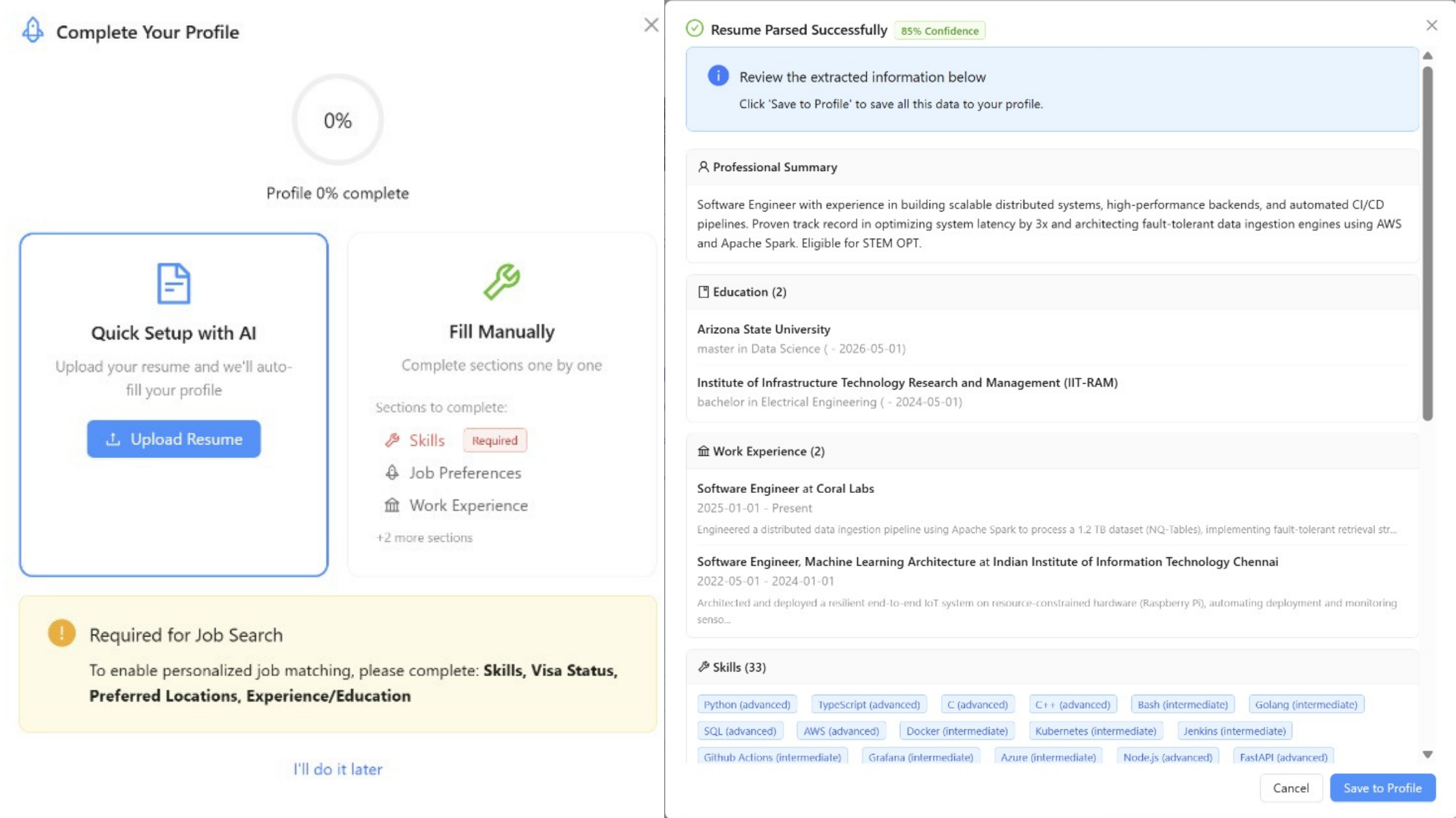}
    \caption{Resume Onboarding}
    \label{fig:resume_onboarding}
  \end{subfigure}
  \hfill
  \begin{subfigure}[b]{0.48\textwidth}
    \centering
    \includegraphics[width=\linewidth, height=0.45\textheight, keepaspectratio]{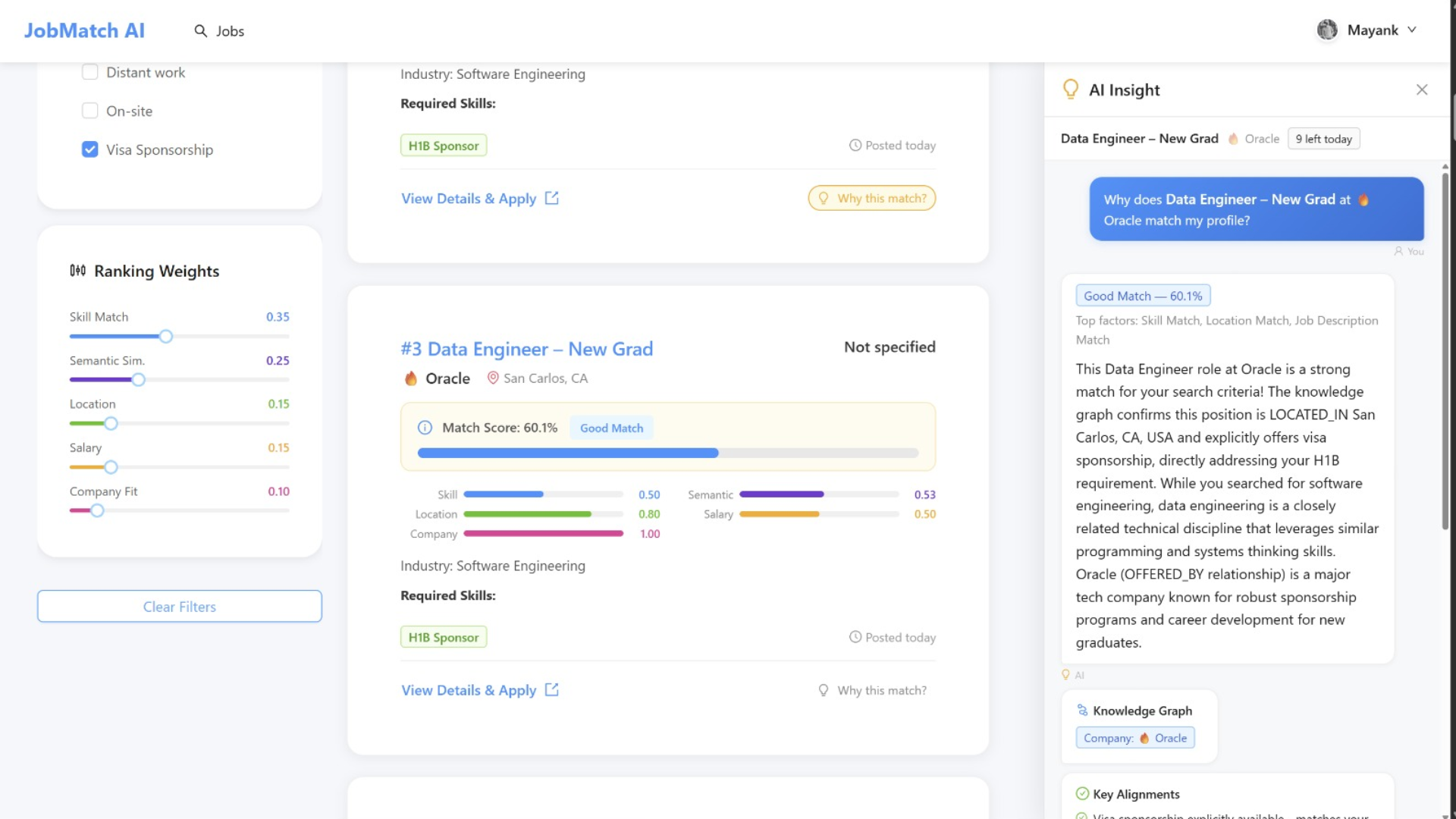}
    \caption{Ranked Results and Explanation}
    \label{fig:ranked_explanation}
  \end{subfigure}
  \caption{JobMatchAI Platform Interface}
  \label{fig:platform_interface}
\end{figure*}

\section{System Demonstration}
JobMatchAI is available as a live web application\footnote{\url{https://mutu.dev}} and as an installable package\footnote{\url{https://github.com/coral-lab-asu/job-hunt-AI.git}}. A screencast video accompanies this submission. The system exposes two interaction modes, each designed to highlight a different aspect of the hybrid retrieval and explainable reranking pipeline.

\paragraph{Smart Search (Resume-Driven).} A user uploads a resume, and the system extracts a structured profile based on skills, experience level, location preferences, and education, which is displayed for verification (Figure~\ref{fig:resume_onboarding}). The hybrid pipeline (Section~\ref{sec:hybrid_pipeline}) then returns jobs ranked by the multi-factor utility score. Each result card shows the overall match percentage alongside a per-factor breakdown (skill fit, experience, location, salary, semantic similarity, company fit). Clicking a result expands an AI explanation panel that narrates the score in natural language, grounded in knowledge-graph paths and factor values (Figure~\ref{fig:ranked_explanation}). To illustrate controllability, users can adjust reranking weight sliders; for instance, prioritizing salary over location, and the list re-orders in real time, with explanations automatically reflecting the updated weights.

\begin{figure}
    \centering
    \includegraphics[width=0.94\linewidth]{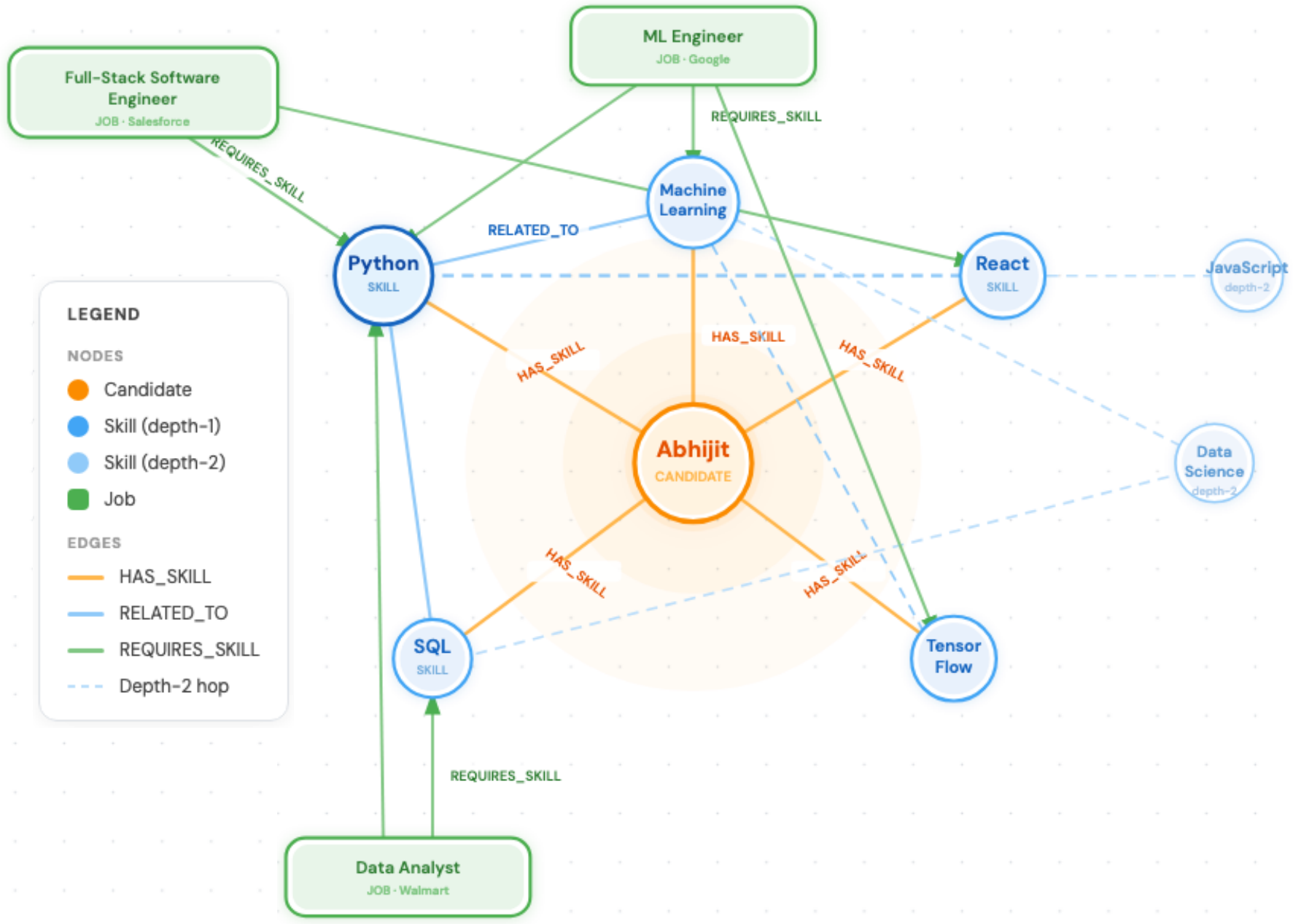}
    \caption{Visualization of the JobMatchAI knowledge graph illustrating the relationships between candidates, jobs, skills, locations, and companies used for hybrid retrieval and explainable reranking.}
    \label{fig:KG_Visualization}
\end{figure}

\paragraph{Keyword Search with Semantic Enrichment.} A free-text query such as \texttt{``junior backend engineer Go"} triggers query enrichment (Section~\ref{sec:hybrid_pipeline}, Stage 1), where graph-based skill expansion associates \texttt{Go} with related concepts such as distributed systems and microservices. Results thus include jobs whose listed requirements differ lexically from the query but align semantically via the knowledge graph. An interactive knowledge-graph neighborhood visualization (Figure~\ref{fig:KG_Visualization}) lets users explore skill clusters and the \texttt{RELATED\_TO} edges that drove retrieval, making the expansion logic transparent rather than hidden.

\section{Evaluation}
\label{sec:evaluation}
We evaluate JobMatchAI along four axes: (i) offline retrieval quality on the JobSearch-XS benchmark with bootstrap confidence intervals (\S\ref{sec:offline}), (ii) per-query variance characterization by query intent (\S\ref{sec:variance}), (iii) a skill-extractor impact study (\S\ref{sec:extractor-impact}), and (iv) a cross-domain transfer spot-check (\S\ref{sec:cross-domain}), followed by a pilot user study (\S\ref{sec:userstudy}). Component-level evaluations (resume parsing, screening models, robustness slices) are reported in Appendix~\ref{app:eval-details}.

\subsection{Offline Evaluation on JobSearch-XS}
\label{sec:offline}

\paragraph{Benchmark.} JobSearch-XS is a skill-grounded benchmark from 1,283 NYC civil-service roles with 30 queries and skill-disjoint train/dev/test splits for zero-shot generalization. It provides \emph{silver} labels (29K pairs from knowledge-graph skill overlap) and \emph{gold} labels (human-verified pairs, annotated by two judges at $\kappa \approx 0.85$). \textbf{All metrics in Table~\ref{tab:retrieval} are computed on gold labels}. Silver labels are derived from skill-graph overlap, partially overlapping with the reranker's skill-match feature. To avoid evaluation circularity, metrics in Table~\ref{tab:retrieval} use independently annotated gold pairs. Silver labels enable development and large-scale diagnostic analysis.

\paragraph{Results.} We report NDCG@\textit{k} (ranking quality, rewarding
relevant results at higher positions), Recall@\textit{k}
(coverage of relevant documents within the top
\textit{k}), and median latency (P50). Table~\ref{tab:retrieval} reports
retrieval and reranking quality. All recall figures
are computed post-fusion (Stage~3), before
hard-constraint filtering (Stage~4).

\begin{table}[htbp]
\centering\small
\caption{Retrieval performance on JobSearch-XS
(evaluated on human-verified gold labels).
\cmark/\xmark{} indicate active retrieval channels.}
\label{tab:retrieval}
\resizebox{\columnwidth}{!}{%
\begin{tabular}{@{}ccc|cccc|c@{}}
\toprule
\textbf{BM25} & \textbf{Sem.} & \textbf{KG}
& \textbf{NDCG@5} & \textbf{NDCG@10}
& \textbf{R@50} & \textbf{R@100}
& \textbf{P50 (ms)} \\
\midrule
\cmark & \xmark & \xmark
  & 0.721 & 0.756 & 0.49 & 0.57 & 79 \\
\xmark & \xmark & \cmark
  & 0.758 & 0.790 & 1.00\textsuperscript{*} & 1.00\textsuperscript{*} & -- \\
\cmark & \cmark & \xmark
  & 0.740 & 0.771 & 0.36 & 0.42 & 80 \\
\cmark & \xmark & \cmark
  & 0.747 & 0.776 & 0.33 & 0.39 & 81 \\
\cmark & \cmark & \cmark
  & 0.750 & 0.780 & 0.29 & 0.35 & 82 \\
\cmark & \cmark & \cmark\textsuperscript{+R}
  & \textbf{0.785} & \textbf{0.810} & 0.29 & 0.35 & 82 \\
\bottomrule
\end{tabular}%
}
\\[4pt]
\raggedright\footnotesize
\textsuperscript{*}Perfect recall on KG-reachable pairs only.
\textsuperscript{+R}With reranker (as discussed in Sec~\ref{sec:reranker}).
\end{table}
The hybrid pipeline with reranking achieves an NDCG@10 of 0.81, a 7\% relative improvement over the BM25 baseline, with median latency under 82\,ms (P95: 86\,ms), well within interactive thresholds. The mean reciprocal rank (MRR) across all queries is 0.76, indicating the first relevant result typically appears in the top two positions.

\paragraph{Statistical reliability.}
Given the size of the gold set, we report 95\% bootstrap confidence intervals (1{,}000 resamples over queries) and assess pairwise significance via paired bootstrap tests (10{,}000 iterations). Confidence intervals on NDCG@10 are wide (typical width $\approx 0.20$) due to the limited gold-query count, and per-query NDCG@10 exhibits substantial variance (std $= 0.30$, IQR $= 0.45$). This variance is concentrated in non-literal query types rather than spread uniformly across system configurations (\S\ref{sec:variance}). Reported metrics in Table~\ref{tab:retrieval} should be interpreted as point estimates with non-trivial uncertainty at this scale; expanding gold annotation is a priority for future releases.

\paragraph{Per-split analysis.} Performance varies across the skill-disjoint splits: train queries achieve NDCG@10 of 0.89, test queries 0.77, and dev queries 0.40. The dev-split gap reflects the challenge of zero-shot skill generalization over a small query set (10 queries per split); we expect this to stabilize as JobSearch-XS is expanded with additional queries and labels.

\paragraph{Recall gap.} The hybrid system's Recall@100 of 0.35 is lower than the BM25 baseline's 0.57. This arises from the current fusion strategy: Reciprocal Rank Fusion (RRF) merges three ranked lists with a union cap of 400 candidates before hard-constraint filtering, which can aggressively prune the candidate pool. The KG-only baseline achieves perfect recall on graph-reachable pairs but does not scale to the full corpus. Improving fusion-stage recall through higher per-channel $k$ values or learned fusion weights is identified as the primary engineering priority for the future release.

\subsection{Variance by Query Intent}
\label{sec:variance}

We partition 15 gold queries by intent type to analyze retrieval quality variation. Table~\ref{tab:variance} shows NDCG@10 distributions for the Hybrid+Reranker configuration.

\begin{table}[htbp]\centering\small\caption{NDCG@10 distribution across gold queries by query intent (Hybrid+Reranker pipeline).}\label{tab:variance}\begin{tabular}{@{}lcccc@{}}\toprule\textbf{Intent type}    & \textbf{n} & \textbf{Mean} & \textbf{Std} & \textbf{IQR} \\\midrule Job title               & 3  & 0.90 & 0.09 & 0.08 \\Natural language        & 6  & 0.74 & 0.34 & 0.39 \\Skill synonym           & 6  & 0.60 & 0.31 & 0.46 \\\midrule\textbf{Overall}        & 15 & 0.72 & 0.30 & 0.45 \\\bottomrule\end{tabular}\end{table}

Literal title queries are robust (mean $= 0.90$, std $= 0.09$), but natural-language and skill-synonym queries cause most variance. Skill-synonym queries are highly variable (mean $= 0.60$, std $= 0.31$), suggesting improvement opportunities in synonym-expansion rather than lexical or semantic retrieval, motivating future work on KG-driven query rewriting and synonym-aware retrieval.

\subsection{Skill Extraction Impact on Retrieval}
\label{sec:extractor-impact}

The resume parser (App.~\ref{app:eval-details}) operates with high precision (0.94) but low recall (0.12), due to concerns about false-positive skills triggering incorrect KG expansion. We test this by comparing two extractor points in the retrieval pipeline:

\begin{itemize} \item \textbf{High-precision} (current): top-1 skill ($P=0.94$, $R=0.12$). \item \textbf{High-recall}: top-3 skills with ESCO synonym expansion ($P \approx 0.65$, $R \approx 0.50$).\end{itemize}

\begin{table}[htbp]\centering\small\caption{End-to-end retrieval quality under two skill-extractor operating points (Hybrid+Reranker).}\label{tab:extractor-impact}\begin{tabular}{@{}lcc@{}}\toprule\textbf{Extractor} & \textbf{NDCG@10} & \textbf{R@100} \\\midrule
High-precision (current) & 0.791 & \textbf{0.336} \\High-recall & \textbf{0.817} & 0.326 \\\midrule
$\Delta$ (recall $-$ precision) & \textbf{+0.026} & $-0.010$ \\\bottomrule\end{tabular}\end{table}

The high-recall configuration improves NDCG@10 by $+2.6$ points, costing $1.0$ point in R@100 (Table~\ref{tab:extractor-impact}). This contradicts our assumption: extractor \emph{recall} constrains downstream retrieval quality, not precision. False-positive skills are down-weighted by the reranker before corrupting the final ranking, while missed skills narrow the candidate's profile. We recommend recall-favoring extraction with downstream filtering and plan to redesign the extractor accordingly.

\subsection{Cross-Domain Transfer}
\label{sec:cross-domain}

Reviewer feedback raised concerns about generalization beyond NYC civil-service postings. We evaluate transfer to private-sector tech roles by drawing 100 jobs from a production job aggregator spanning software engineering, ML, and data infrastructure roles across companies. We construct 5 representative queries (senior backend, frontend React, ML/recommendations, data engineer, DevOps) and use a weak relevance criterion: a job is relevant if $\geq 2$ query skills appear as whole-word matches in its text. With no domain adaptation, the system achieves macro-averaged NDCG@10 $\approx 0.54$ on this set.

This is a directional rather than rigorous evaluation---the corpus is small, labels are heuristic, and there is no human-annotated gold---but it indicates that the system transfers meaningfully to less-standardized postings. The absolute drop from in-domain quality (NDCG@10 $= 0.81$ on JobSearch-XS) reflects a non-trivial domain shift and motivates future work on domain-adaptive variants of the skill graph and embedding model.

\subsection{Pilot User Study}
\label{sec:userstudy}

To assess real-world usability, we conducted a pilot study with $N=20$ participants recruited from graduate students at various University / job seeker communities. Participants included 15 active job seekers and 5 employed professionals, with experience levels ranging from 0 to 7 years.

\paragraph{Protocol.} Each participant completed four tasks: (1) upload a resume and rate the top-5 Smart Search results for relevance, (2) perform a keyword search and assess whether skill expansion surfaced useful results, (3) read AI-generated match explanations and rate their helpfulness, and (4) adjust reranking weight sliders and evaluate responsiveness. All ratings used a 5-point Likert scale.

\paragraph{Results.} Table~\ref{tab:userstudy} summarizes task-based and usability ratings.
\begin{table}[htbp]
\centering\small
\caption{User study results ($N=20$). Scores are mean $\pm$ std on a 1--5 Likert scale.}
\label{tab:userstudy}
\begin{tabular}{@{}lc@{}}
\toprule
\textbf{Metric} & \textbf{Mean $\pm$ Std} \\
\midrule
\multicolumn{2}{@{}l}{\emph{Task-based evaluation}} \\
Top-5 relevance (Smart Search)  & 3.92 $\pm$ 0.79 \\
Skill synonym handling          & 3.75 $\pm$ 0.87 \\
Explanation helpfulness         & 4.17 $\pm$ 0.72 \\
Weight adjustment usefulness   & 3.83 $\pm$ 0.94  \\
\midrule
\multicolumn{2}{@{}l}{\emph{Usability}} \\
Ease of use                    & 4.08 $\pm$ 0.67 \\
Would use for actual job search & 3.58 $\pm$ 1.00 \\
Acceptable response time       & 4.33 $\pm$ 0.65  \\
\bottomrule
\end{tabular}
\end{table}

Participants identified ``AI match explanations'' and ``adjustable ranking weights'' as the most valuable features. Qualitative feedback highlighted ``appreciation for transparency in scoring'' and ``requests for more granular skill filtering''. We report these ratings as descriptive evidence rather than confirmatory; the absence of a controlled baseline and the homogeneous participant pool (no recruiters) limit the strength of conclusions that can be drawn.

\paragraph{Explanation faithfulness.} To assess explanation accuracy, we audited 95 system-generated explanations across 16 user profiles against three criteria: (C1) whether explanations mention the highest-contributing factor, (C2) whether factors below 0.5 are shown as weaknesses, and (C3) whether explanations avoid unsupported claims.

\textbf{Results:} The top factor was mentioned in 67/95 cases (70.5\%); among the 91 explanations where at least one factor scored below 0.5, the weakness was surfaced in 86 cases (94.5\%); and no unsupported claims were found in any of the 95 explanations (100\%). C1 performance varied: 93.8\% for high-match versus 58--59\% for medium and low-match results. C1 failures primarily showed \emph{omission}: for weaker matches, the LLM focuses on explaining misalignments over stating the highest-scoring factor. This aligns with architectural constraints, i.e., the model narrates existing scores without inventing evidence. The C1 criterion uses keyword matching, which may undercount implicit references; future work will use human or LLM judges for more rigorous evaluation.

\subsection{Limitations}
\label{sec:limitations}
JobSearch-XS uses only 30 queries and 40 gold-labeled pairs from NYC civil-service postings, limiting domain coverage and metric reliability. The hybrid system's Recall@100 of 0.35 falls below the BM25 baseline due to precision-focused fusion. The knowledge graph relies on a curated ESCO synonym table, missing emerging skills. The pilot study (N=20) lacks statistical power. While the utility function is auditable, no bias audit exists for non-traditional candidates.
\section{Related Work}
\paragraph{Semantic retrieval and knowledge graphs.} Dense retrieval models such as Sentence-BERT \cite{reimers-2019-sentence-bert} and ColBERT \cite{khattab_colbert_2020,santhanam_colbertv2_2022} have advanced semantic matching on benchmarks like MS MARCO \cite{bajaj_ms_2018}, but typically lack integration with domain-specific constraints. Knowledge-graph-aware methods such as KGAT \cite{wang_kgat_2019} combine graph structure with learned representations for recommendation tasks, yet rarely expose interpretable, per-factor utility features. JobMatchAI bridges both lines of work by fusing dense retrieval with a typed skill knowledge graph and feeding graph-derived features into a transparent reranker.
\newline\indent\textbf{Explainability in hiring.} High-stakes domains such as hiring increasingly require structurally grounded explanations rather than post-hoc attention visualizations or feature-importance approximations \cite{ribeiro_why_2016,jain_attention_2019}, to support audit logging and regulatory compliance \cite{doshi-velez_towards_2017,raghavan_mitigating_2020}. JobMatchAI addresses this by strictly separating a deterministic, white-box scoring layer from an optional generative explanation layer, ensuring that every ranking decision is mathematically traceable.
\newline\indent\textbf{Job matching systems.} Commercial platforms such as LinkedIn, Jobright, Teal, and Jobscan rely predominantly on keyword search with proprietary heuristics and offer limited transparency into ranking logic. General-purpose retrieval frameworks like PyTerrier~\cite{macdonald_declarative_2020} and Vespa~\cite{bergum_e-commerce_2019} provide strong retrieval infrastructure but do not encode hiring-specific constraints. Table~\ref{tab:comparison} contrasts JobMatchAI with representative systems across six capabilities; to our knowledge, no existing public system combines hybrid retrieval, knowledge-graph grounding, explainable factor-level scoring, and a released benchmark.
\begin{table}[t]
\centering\small
\setlength{\tabcolsep}{3pt}
\caption{Feature comparison with representative job search tools. \cmark = supported, \xmark = not supported, \textasciitilde = partial.}\label{tab:comparison}
\begin{tabular}{@{}lccccc@{}}
\toprule
\textbf{Feature} & \rotatebox{60}{\textbf{JobMatchAI}} & \rotatebox{60}{\textbf{LinkedIn}} & \rotatebox{60}{\textbf{Jobright}} & \rotatebox{60}{\textbf{Teal}} & \rotatebox{60}{\textbf{Jobscan}} \\
\midrule
Semantic search        & \cmark & \textasciitilde & \cmark & \xmark & \xmark \\
Knowledge graph        & \cmark & \xmark & \xmark & \xmark & \xmark \\
Hybrid retrieval       & \cmark & \xmark & \xmark & \xmark & \xmark \\
Explainable scores     & \cmark & \xmark & \xmark & \xmark & \textasciitilde \\
Factor-level control   & \cmark & \xmark & \xmark & \xmark & \xmark \\
Resume-driven search   & \cmark & \textasciitilde & \cmark & \cmark & \cmark \\
Public benchmark       & \cmark & \xmark & \xmark & \xmark & \xmark \\
\bottomrule
\end{tabular}
\\[4pt]
$\dagger$ \raggedright\scriptsize {Capability assessments based on publicly available documentation, independent published analyses, and hands-on testing as of February 2026. Capabilities of commercial products may have changed since.}
\end{table}

\section{Conclusion}
JobMatchAI combines Sentence Transformer embeddings, a skill knowledge graph, and BM25 indexing for explainable job search. The system separates deterministic scoring from LLM explanations for traceable rankings, and achieves NDCG@10 of 0.81 at sub-100\,ms latency on JobSearch-XS. We characterize the system's behavior beyond a single headline number: bootstrap confidence intervals quantify metric uncertainty, per-query variance analysis localizes weakness to non-literal queries (\S\ref{sec:variance}), a skill-extractor ablation overturns the original precision-focused design (\S\ref{sec:extractor-impact}), and a cross-domain spot-check provides directional evidence of transfer (\S\ref{sec:cross-domain}). Future work will improve fusion-stage recall, strengthen synonym-driven query rewriting, and scale JobSearch-XS with denser gold annotation. We release the system\footnote{Code is available under the MIT License at \url{https://github.com/coral-lab-asu/job-hunt-AI}. The JobSearch-XS benchmark released under CC-BY-4.0.} for explainable architectures.

\section*{Ethics and Broader Impact}
Hiring is a domain with significant ethical and legal implications. JobMatchAI is designed as decision-support tooling, it surfaces and explains ranked results but does not make autonomous hiring decisions. Several design choices reflect this intent: the white-box utility function enables per-factor audit logging, the weight vector is user-adjustable rather than fixed, and the LLM explanation layer is downstream of deterministic scoring to prevent hallucinated justifications.
\newline\indent Nonetheless, risks remain. Embedding models and knowledge-graph structures may encode historical biases, potentially under-weighting non-traditional career paths or under-represented skill vocabularies. Match scores, though transparent in construction, could be misinterpreted as objective measures of candidate worth rather than as relevance estimates conditioned on a specific job description. To mitigate these risks, we recommend that deployments (i) treat rankings as one input among many in hiring workflows, (ii) incorporate fairness-aware evaluation metrics into the benchmark harness, and (iii) conduct domain-specific bias audits before production use, particularly examining whether skill-graph topology disadvantages candidates from non-traditional educational or career backgrounds.

\bibliography{references,custom}

@inproceedings{reimers-2019-sentence-bert,
  title     = {Sentence-BERT: Sentence Embeddings using Siamese BERT-Networks},
  author    = {Reimers, Nils and Gurevych, Iryna},
  booktitle = {Proceedings of the 2019 Conference on Empirical Methods in Natural Language Processing},
  year      = {2019}
}

@misc{doshi-velez_towards_2017,
	title = {Towards {A} {Rigorous} {Science} of {Interpretable} {Machine} {Learning}},
	url = {http://arxiv.org/abs/1702.08608},
	doi = {10.48550/arXiv.1702.08608},
	urldate = {2025-12-02},
	publisher = {arXiv},
	author = {Doshi-Velez, Finale and Kim, Been},
	month = mar,
	year = {2017},
	note = {arXiv:1702.08608 [stat]},
}

@misc{jain_attention_2019,
	title = {Attention is not {Explanation}},
	url = {http://arxiv.org/abs/1902.10186},
	doi = {10.48550/arXiv.1902.10186},
	urldate = {2025-12-02},
	publisher = {arXiv},
	author = {Jain, Sarthak and Wallace, Byron C.},
	month = may,
	year = {2019},
	note = {arXiv:1902.10186 [cs]},
}

@misc{bergum_e-commerce_2019,
	title = {E-{Commerce} {Search} and {Recommendation} with {Vespa}.ai},
	url = {https://medium.com/vespa/e-commerce-search-and-recommendation-with-vespa-ai-a49c98f97e68},
	language = {en},
	urldate = {2025-12-02},
	journal = {vespa},
	author = {Bergum, Jo Kristian},
	month = {nov},
	year = {2019},
}

@inproceedings{macdonald_declarative_2020,
	title = {Declarative {Experimentation} in {Information} {Retrieval} using {PyTerrier}},
	url = {http://arxiv.org/abs/2007.14271},
	doi = {10.1145/3409256.3409829},
	urldate = {2025-12-02},
	booktitle = {Proceedings of the 2020 {ACM} {SIGIR} on {International} {Conference} on {Theory} of {Information} {Retrieval}},
	author = {Macdonald, Craig and Tonellotto, Nicola},
	month = sep,
	year = {2020},
	note = {arXiv:2007.14271 [cs]},
	pages = {161--168},
}

@inproceedings{raghavan_mitigating_2020,
	title = {Mitigating {Bias} in {Algorithmic} {Hiring}: {Evaluating} {Claims} and {Practices}},
	shorttitle = {Mitigating {Bias} in {Algorithmic} {Hiring}},
	url = {http://arxiv.org/abs/1906.09208},
	doi = {10.1145/3351095.3372828},
	urldate = {2025-12-02},
	booktitle = {Proceedings of the 2020 {Conference} on {Fairness}, {Accountability}, and {Transparency}},
	author = {Raghavan, Manish and Barocas, Solon and Kleinberg, Jon and Levy, Karen},
	month = jan,
	year = {2020},
	note = {arXiv:1906.09208 [cs]},
	pages = {469--481},
}

@misc{ribeiro_why_2016,
	title = {"{Why} {Should} {I} {Trust} {You}?": {Explaining} the {Predictions} of {Any} {Classifier}},
	shorttitle = {"{Why} {Should} {I} {Trust} {You}?},
	url = {http://arxiv.org/abs/1602.04938},
	doi = {10.48550/arXiv.1602.04938},
	urldate = {2025-12-02},
	publisher = {arXiv},
	author = {Ribeiro, Marco Tulio and Singh, Sameer and Guestrin, Carlos},
	month = aug,
	year = {2016},
	note = {arXiv:1602.04938 [cs]},
}

@misc{bajaj_ms_2018,
	title = {{MS} {MARCO}: {A} {Human} {Generated} {MAchine} {Reading} {COmprehension} {Dataset}},
	shorttitle = {{MS} {MARCO}},
	url = {http://arxiv.org/abs/1611.09268},
	doi = {10.48550/arXiv.1611.09268},
	urldate = {2025-12-02},
	publisher = {arXiv},
	author = {Bajaj, Payal and Campos, Daniel and Craswell, Nick and Deng, Li and Gao, Jianfeng and Liu, Xiaodong and Majumder, Rangan and McNamara, Andrew and Mitra, Bhaskar and Nguyen, Tri and Rosenberg, Mir and Song, Xia and Stoica, Alina and Tiwary, Saurabh and Wang, Tong},
	month = oct,
	year = {2018},
	note = {arXiv:1611.09268 [cs]},
}

@misc{khattab_colbert_2020,
	title = {{ColBERT}: {Efficient} and {Effective} {Passage} {Search} via {Contextualized} {Late} {Interaction} over {BERT}},
	shorttitle = {{ColBERT}},
	url = {http://arxiv.org/abs/2004.12832},
	doi = {10.48550/arXiv.2004.12832},
	urldate = {2025-11-11},
	publisher = {arXiv},
	author = {Khattab, Omar and Zaharia, Matei},
	month = jun,
	year = {2020},
	note = {arXiv:2004.12832 [cs]},
}

@inproceedings{wang_kgat_2019,
	address = {New York, NY, USA},
	series = {{KDD} '19},
	title = {{KGAT}: {Knowledge} {Graph} {Attention} {Network} for {Recommendation}},
	isbn = {978-1-4503-6201-6},
	shorttitle = {{KGAT}},
	url = {https://doi.org/10.1145/3292500.3330989},
	doi = {10.1145/3292500.3330989},
	urldate = {2025-11-10},
	booktitle = {Proceedings of the 25th {ACM} {SIGKDD} {International} {Conference} on {Knowledge} {Discovery} \& {Data} {Mining}},
	publisher = {Association for Computing Machinery},
	author = {Wang, Xiang and He, Xiangnan and Cao, Yixin and Liu, Meng and Chua, Tat-Seng},
	month = jul,
	year = {2019},
	pages = {950--958},
}

@misc{santhanam_colbertv2_2022,
	title = {{ColBERTv2}: {Effective} and {Efficient} {Retrieval} via {Lightweight} {Late} {Interaction}},
	shorttitle = {{ColBERTv2}},
	url = {http://arxiv.org/abs/2112.01488},
	doi = {10.48550/arXiv.2112.01488},
	urldate = {2025-11-11},
	publisher = {arXiv},
	author = {Santhanam, Keshav and Khattab, Omar and Saad-Falcon, Jon and Potts, Christopher and Zaharia, Matei},
	month = jul,
	year = {2022},
	note = {arXiv:2112.01488 [cs]},
}
\newpage
\appendix
\section{Extended Evaluation Details}
\label{app:eval-details}
\subsection{JobSearch-XS Benchmark Construction} JobSearch-XS evaluates three key capabilities distinguishing hybrid job-matching systems from standard retrieval: (i) skill-synonym resolution, (ii) structured metadata matching (location, salary, seniority), and (iii) zero-shot generalization to unseen skill vocabularies. The construction pipeline is as follows.

\textbf{Source data.} We use a fixed snapshot of the ``NYC Jobs'' dataset from NYC Open Data (Socrata endpoint \texttt{kpav-sd4t}), containing 1,283 civil-service job postings. Each includes title, description, salary range, location, required and preferred skills, providing structured metadata for multi-factor evaluation. This snapshot is archived with release artifacts for reproducibility.

\textbf{Skill normalization.} Raw skill mentions are canonicalized using ESCO v1.1.1 plus a manually-curated synonym table mapping surface forms to canonical skill IDs (e.g., ``k8s'' $\to$ Kubernetes, ``ML'' $\to$ Machine Learning). This enables consistent skill-overlap computation and forms the vocabulary for knowledge-graph population.

\textbf{Query generation.} Thirty queries are generated across three templates:
\begin{itemize}
    \item \textbf{Title-based} (10 queries): From job titles (e.g., ``Data Analyst''), testing retrieval of matching and related roles.
    \item \textbf{Natural-language} (10 queries): Paraphrased role descriptions (e.g., ``entry-level position analyzing public health data''), testing semantic understanding beyond keywords.
    \item \textbf{Skill-synonym} (10 queries): Using skill synonyms or related terms absent from corpus vocabulary (e.g., ``container orchestration'' for ``Docker'' and ``Kubernetes''), targeting knowledge-graph expansion and embedding generalization.
\end{itemize}

\textbf{Label generation.} Silver relevance labels (~29K query–document pairs) are generated by computing skill overlap via the knowledge graph. A pair is positive if the Jaccard similarity of canonical skill sets exceeds threshold $\tau=0.3$, including depth-2 graph-expanded skills. A small gold set is manually annotated: for each dev and test query, one relevant and one irrelevant job verified by two annotators, with inter-annotator agreement $\kappa \approx 0.85$ (Cohen's kappa).

\textbf{Skill-disjoint splits.} The 30 queries are split into train (10), dev (10), and test (10) so canonical skill sets are maximally disjoint. A greedy set-cover assignment minimizes skill overlap, leaving ~60\% of test-split skills unseen in training. This forces generalization beyond memorized skill co-occurrences, vital for real-world deployment with emerging skills and job categories.

\textbf{Build pipeline.} A reproducible script executes: (1) loading NYC Jobs snapshot, (2) applying ESCO-based skill normalization, (3) running semantic ingestion (embedding and entity extraction), (4) generating templated queries, (5) computing silver labels via graph skill overlap, and (6) outputting benchmark file, summary manifest, and data slices. The pipeline completes in under 10 minutes on a single machine.

\textbf{Limitations.} JobSearch-XS is intentionally small—30 queries over 1,283 documents—and from a single domain (NYC civil-service roles). The silver labeling may introduce false positives from coincidental skill overlap with different contexts. The gold annotation set is minimal (2 labels per dev/test query). JobSearch-XS serves as a starter benchmark demonstrating evaluation methodology; scaling to larger, multi-domain benchmarks with denser gold annotations is a key future goal.
\subsection{Resume Parsing}
The resume parser uses regex, positional heuristics, and NER for name extraction, combined with a hybrid skill extractor using semantic similarity and spaCy patterns, evaluated on 45 labeled resumes across three roles.
Table~\ref{tab:resume-parsing} reports macro-averaged F1, with precision and recall reported for skills and education.

\begin{table}[ht]
\centering
\small
\caption{Component-level resume parsing quality on a labeled set of 45 resumes. Baseline uses a single-strategy spaCy NER pipeline.}
\resizebox{\columnwidth}{!}{
\begin{tabular}{@{}lccccccr@{}}
\toprule
\textbf{Field} & \textbf{Base P} & \textbf{Base R} & \textbf{Base F1}
& \textbf{Ours P} & \textbf{Ours R} & \textbf{Ours F1} & \textbf{$\Delta$F1} \\
\midrule
Name       & --   & --   & 0.33 & --   & --   & 0.64 & +0.31 \\
Skills     & 0.41 & 0.60 & 0.48 & 0.94 & 0.12 & 0.19 & $-$0.29 \\
Education  & 0.74 & 0.31 & 0.42 & 1.00 & 1.00 & 1.00 & +0.58 \\
\bottomrule
\end{tabular}
}
\label{tab:resume-parsing}
\end{table}



\subsection{Resume Screening Models}
Name extraction improves by +0.31 F1 using positional heuristics on non-standard headers. Education parsing achieves perfect precision and recall (+0.58 F1) via pattern matching. Skill extraction favors precision (0.94) over recall (0.12), lowering F1 by -0.29.However, the end-to-end ablation in \S\ref{sec:extractor-impact} reveals extractor \emph{recall} as the key constraint for downstream retrieval quality. Recall limits retrieval quality; extractor redesign is planned.


\subsection{Resume Screening Models} JobMatchAI’s pairwise screening models, trained on 1,200 candidate–job triples, serve as baselines. The pairwise MLP achieves 0.70 accuracy and 0.72 AUROC, while the off-the-shelf cross-encoder performs poorly (0.30 accuracy, 0.51 AUROC). Domain-specific fine-tuning is planned for improvement.
Table~\ref{tab:screening} reports performance.

\begin{table}[h]\centering\small\caption{Resume screening model performance on pairwise preference prediction (1,200 triples, 80/20 train/test split).}\label{tab:screening}\resizebox{\columnwidth}{!}{\begin{tabular}{@{}lcccp{3.8cm}@{}}\toprule\textbf{Model} & \textbf{Acc.} & \textbf{F1} & \textbf{AUROC} & \textbf{Notes} \\\midrule Pairwise MLP & 0.70 & 0.45 & 0.72 & Concatenated embedding features; reasonable baseline \\ Cross-encoder & 0.30 & 0.34 & 0.51 & Off-the-shelf without fine-tuning; poor calibration \\\bottomrule\end{tabular}}\end{table}

\subsection{Robustness Slices}
To analyze retrieval quality across structured dimensions, JobSearch-XS queries are partitioned by location match, salary alignment, and target seniority. Table~\ref{tab:slices} shows NDCG@10 and Recall@100 for the hybrid + reranker setup across these slices.

\begin{table}[h]\centering\caption{Retrieval performance by location, salary, and seniority slices on JobSearch-XS (hybrid + reranker).}\label{tab:slices}\resizebox{\columnwidth}{!}{\begin{tabular}{@{}llccl@{}}\toprule\textbf{Dimension} & \textbf{Slice} & \textbf{NDCG@10}& \textbf{Recall@100} & \textbf{Notes} \\\midrule Location  & Same-region      & 0.78 & 0.21 & Best precision with location match \\Location  & Cross-region     & 0.50 & 0.85 & High recall, low ranking precision \\Location  & Remote-friendly  & 0.66 & 0.57 & Moderate precision and recall \\\midrule Salary    & Within band      & 0.77 & 0.23 & Highest NDCG with salary overlap \\Salary    & Unknown/noisy    & 0.63 & 1.00 & Perfect recall; fallback when salary missing \\\midrule Seniority & Junior           & 0.62 & 0.92 & High recall, lower ranking quality \\Seniority & Mid-level        & 0.67 & 1.00 & Perfect recall; balanced performance \\Seniority & Senior           & 0.69 & 0.23 & Good precision, low recall due to complexity \\\bottomrule\end{tabular}}\end{table}

Results show precision-recall tradeoffs. Same-region and within-salary-band queries have highest NDCG@10 (0.78, 0.77) with strong ranking, but lower recall (0.21–0.23). Cross-region, unknown salary, and junior/mid-level slices have high recall (0.85–1.00) but less ranking precision. Senior-level queries show decent NDCG@10 (0.69) but low recall (0.23) due to complex factors. These patterns suggest structured features improve ranking, while missing data shifts focus to recall, motivating future feature imputation research.

\subsection{Auxiliary Statistics}

\FloatBarrier
\begin{table}[!htbp]
\centering\tiny
\caption{JobSearch-XS dataset statistics.}
\label{tab:dataset}
\begin{tabular}{@{}lr@{}}
\toprule
\textbf{Statistic} & \textbf{Value} \\
\midrule
Total Queries  & 30 \\
Total Jobs     & 1{,}283 \\
Silver Labels  & 29{,}013 \\
Gold Labels    & 30 \\
\midrule
Train / Dev / Test & 15 / 6 / 9 \\
\bottomrule
\end{tabular}
\end{table}
\vspace{-20em}
\begin{table}[h]
\centering\tiny
\caption{Latency profile (210 queries, 12.24 qps).}
\label{tab:latency}
\begin{tabular}{@{}lcccc@{}}
\toprule
& \textbf{P50} & \textbf{P90} & \textbf{P95} & \textbf{Mean $\pm$ Std} \\
\midrule
Latency (ms) & 81.55 & 84.66 & 86.00 & 81.67 $\pm$ 2.69 \\
\bottomrule
\end{tabular}
\end{table}
\FloatBarrier
\begin{table}[!htbp]
\centering\tiny
\caption{\small Information retrieval metrics on JobSearch-XS (baseline).}
\label{tab:ir-metrics}
\begin{tabular}{@{}lcccccc@{}}
\toprule
\textbf{Metric} & \textbf{@1} & \textbf{@3} & \textbf{@5}
  & \textbf{@10} & \textbf{@20} & \textbf{@50} \\
\midrule
NDCG      & 0.667 & 0.720 & 0.733 & 0.756 & 0.759 & 0.772 \\
Recall    & 0.010 & 0.037 & 0.061 & 0.131 & 0.262 & 0.574 \\
Precision & 0.667 & 0.733 & 0.747 & 0.773 & 0.767 & 0.703 \\
\bottomrule
\end{tabular}
\\[4pt]
\raggedright\scriptsize
{MRR\,=\,0.764 \quad MAP\,=\,0.753}
\end{table}

\FloatBarrier
\begin{table}[!htbp]
\centering\tiny
\caption{Per-split retrieval results.}
\label{tab:per-split}
\begin{tabular}{@{}lcccc@{}}
\toprule
\textbf{Split} & \textbf{Queries} & \textbf{NDCG@10}
  & \textbf{MRR} & \textbf{P50 (ms)} \\
\midrule
Train & 15 & 0.890 & 0.922 & 79.08 \\
Dev   &  6 & 0.402 & 0.389 & 77.23 \\
Test  &  9 & 0.769 & 0.749 & 78.23 \\
\bottomrule
\end{tabular}
\end{table}

\FloatBarrier
\begin{table}[!htbp]
\centering\scriptsize
\caption{Per-profile personalized search results.}
\label{tab:per-profile}
\begin{tabular}{@{}lcccc@{}}
\toprule
\textbf{Profile} & \textbf{NDCG@10} & \textbf{R@10}
  & \textbf{MRR} & \textbf{P50 (ms)} \\
\midrule
Entry SWE        & 0.721 & 0.144 & 0.780 & 81.90 \\
Senior Data Eng  & 0.642 & 0.148 & 0.747 & 82.32 \\
H1B ML           & 0.642 & 0.148 & 0.747 & 83.10 \\
Public Health    & 0.721 & 0.144 & 0.780 & 82.87 \\
GIS Analyst      & 0.721 & 0.144 & 0.780 & 82.88 \\
\bottomrule
\end{tabular}
\end{table}

\end{document}